\documentclass[a4paper]{article}
\usepackage{INTERSPEECH2018}

\usepackage{multicol}
\usepackage{graphicx}
\usepackage{multirow}
\usepackage{amssymb,amsmath,bm}
\usepackage{enumitem}
\usepackage{color}
\usepackage{array}
\usepackage{footnote}
\makesavenoteenv{table}
\makesavenoteenv{algorithm}
\usepackage{algorithm}
\usepackage[noend]{algpseudocode}
\usepackage{hyperref}
\usepackage{color}

\usepackage{listings}             

\title{A GPU-based WFST Decoder with Exact Lattice Generation }
\name{Zhehuai Chen$^{1,3}$, Justin Luitjens$^2$, Hainan Xu$^3$, Yiming Wang$^3$, Daniel Povey$^3$, Sanjeev Khudanpur$^3$
\thanks{
This work was partially supported by DARPA LORELEI Grant No
HR0011-15-2-0024 and NSF Grant No CRI-1513128. 
The first author was supported by the National Key Research and Development Program of China under Grant No.2017YFB1002102, the China NSFC project (No. 61573241) and the Shanghai International Science and Technology Cooperation Fund (No. 16550720300).
}
}
\address{
  $^1$SpeechLab, Department of Computer Science and Engineering, Shanghai Jiao Tong University\\
  $^2$NVIDIA, USA\\
  $^3$Center for Language and Speech Processing, Johns Hopkins University}
\email{chenzhehuai@sjtu.edu.cn, jluitjens@nvidia.com, dpovey@gmail.com, \{hxu31,yiming.wang,khudanpur\}@jhu.edu}

\begin{document}


\definecolor{mygreen}{rgb}{0,0.5,0}
\definecolor{mygray}{rgb}{0.5,0.5,0.5}
\definecolor{mymauve}{rgb}{0.58,0,0.82}
\lstset{ %
backgroundcolor=\color{white},   
basicstyle=\footnotesize\ttfamily,        
columns=fullflexible,
breaklines=true,                 
captionpos=b,                    
tabsize=4,
commentstyle=\color{mygreen},    
escapeinside={\%*}{*)},          
keywordstyle=\color{blue},       
stringstyle=\color{mymauve}\ttfamily,     
frame=single,
language=c++,
}

\maketitle
\begin{abstract}
\vspace{-0.25em} 

We describe initial work on an extension of the Kaldi toolkit that supports weighted finite-state
transducer (WFST) decoding on Graphics Processing Units (GPUs).
We implement token recombination as an atomic GPU operation in order to fully parallelize
the Viterbi beam search, and propose a dynamic load balancing strategy for more efficient token
passing scheduling among GPU threads.
We also redesign the exact lattice generation and lattice pruning algorithms for better utilization
of the GPUs. Experiments on the Switchboard corpus show that the proposed method achieves identical  1-best results and lattice quality in recognition and confidence measure tasks, while running 3 to 15 times faster than the  single process Kaldi decoder. The above results are reported on different GPU architectures.
Additionally we obtain a 46-fold speedup with sequence parallelism and multi-process service (MPS) in GPU.

\end{abstract}

\vspace{-0.25em}
\section{Introduction}
\label{sec:intro}
Recent advances in deep learning based automatic speech recognition (ASR) invoke
growing demands in large scale speech transcription and analysis. The 
main computation of a common speech transcription system includes:
acoustic model (AM) inference and weighted finite-state transducer (WFST) decoding. 
%
To reduce computation of the AM inference, researchers have proposed a variety of efficient forms of AMs,
including novel structures~\cite{xue2014singular,peddinti2018low}, quantization
~\cite{mcgraw2016personalized}, frame-skipping~\cite{pundak2016lower,zhc00-chen-is16,zhc00-chen-tasl2017}
and end-to-end systems~\cite{audhkhasi2017direct,e2e-2018}.
Meanwhile, algorithmic improvement, e.g. pruning~\cite{mohri2002weighted,hori2004fast}
and lookahead~\cite{soltau2009dynamic,nolden2012search} is the common way
to speed up decoding.

GPU-based parallel computing is another potential direction which utilizes a large number of  units to parallelize the computation.
As most of the computation takes the form of matrix operation, 
the AM inference can be sped up by sequence batching~\cite{dixon2009harnessing} similar to that during  training~\cite{vesely2010parallel}.
Nevertheless,  GPU-based decoding is less prevalent,
despite its success in small language model (LM)~\cite{you2009parallel}.
Two reasons hamper its wide application: i) it is hard to utilize large LMs since GPUs have smaller memory. 
ii) the solution is not general, e.g. with specific AM or LM requirements,  and lack of lattice generation. 
%

This work is an extension to  Kaldi~\cite{povey2011kaldi},
applying GPU parallel computing in WFST decoding. 
It is a general-purpose offline decoder~\footnote{Recent advances in CPU decoding make it easier to build a real-time decoder
 ~\cite{peddinti2018low,zhc00-chen-tasl2017} for online streaming applications.
 Thus our goal is to transcribe tons of offline audios.},
which does not impose special requirements on the form of AM or LM, and works for
all GPU architectures.
To utilize large LMs in the 2nd pass and support rich post-processing, our design is to decode
the WFSTs and generate exact lattices~\cite{povey2012generating}.
The work is open-sourced~\footnote{\url{https://github.com/chenzhehuai/kaldi/tree/gpu-decoder}},
and will be compatible with all released Kaldi recipes.


The rest of the paper is organized as follows. Prior work is compared in Section~\ref{sec:relate}. 
Our main contributions are described in Sections~\ref{sec:viterbi-search}:
i)  implement token recombination as an atomic GPU operation with no precision loss.
ii)  propose a load balancing strategy for token passing scheduling among GPU threads.
iii) redesign parallel versions of exact lattice generation and pruning.
Experiments are conducted on the Switchboard corpus in Section~\ref{sec:exp},
followed by  conclusion in Section~\ref{sec:conclu}.

\vspace{-0.25em}
\section{Comparisons with Prior Work}
\label{sec:relate}

\cite{mendis2016parallelizing} introduced a parallel Viterbi algorithm in CPU,
which concentrates on partitioning the WFSTs to trade off
between load balancing and synchronization overhead.
To cope with similar issues
in the context of GPU architectures, We give  general LM-independent
solutions in Section~\ref{sec:para-viterbi}. 
\cite{you2009parallel,chong2010exploring} first utilized GPUs in ASR decoding
and showed significant speedup.
However,  the limited GPU memory
becomes a bottleneck for large vocabulary tasks.
A series of works were proposed in~\cite{kim2011h,kim2012efficient,kim2014accelerating}
and attempted to cope with large LM issue by an on-the-fly rescoring~\cite{hori2004fast} algorithm.
Advantages of our method include: i)  general solution and better performance, by generating an exact lattice in the first pass decoding followed by a second pass rescoring, which always obtains slightly better results~\cite{sak2010fly}. The resultant lattice also benefits other speech processing tasks, e.g. confidence measure. 
ii) faster decoding speed. This work benefits from less synchronization overheads, better load balancing, and sequence parallelism. Moreover, it also requires less data transfer compared with~\cite{kim2014accelerating}.
Finally, to the best of our understanding, \cite{argueta2017decoding} is the only open-source project in this field, but it only implemented the basic Viterbi decoding without combining AM posteriors and beam~\cite{mohri2002weighted}, which cannot be applied to ASR. 

\vspace{-0.25em}
\section{Parallel Viterbi Beam Search}
\label{sec:viterbi-search}
The proposed system works in a 2-pass decoding framework~\cite{woodland19951994}
to tackle the language model size problem and enable rich post-processing based on exact lattices.
\vspace{-0.5em}
\subsection{System Outline}
\vspace{-0.5em}
\begin{figure*}[ht]
  \centering
    \includegraphics[width=1\linewidth]{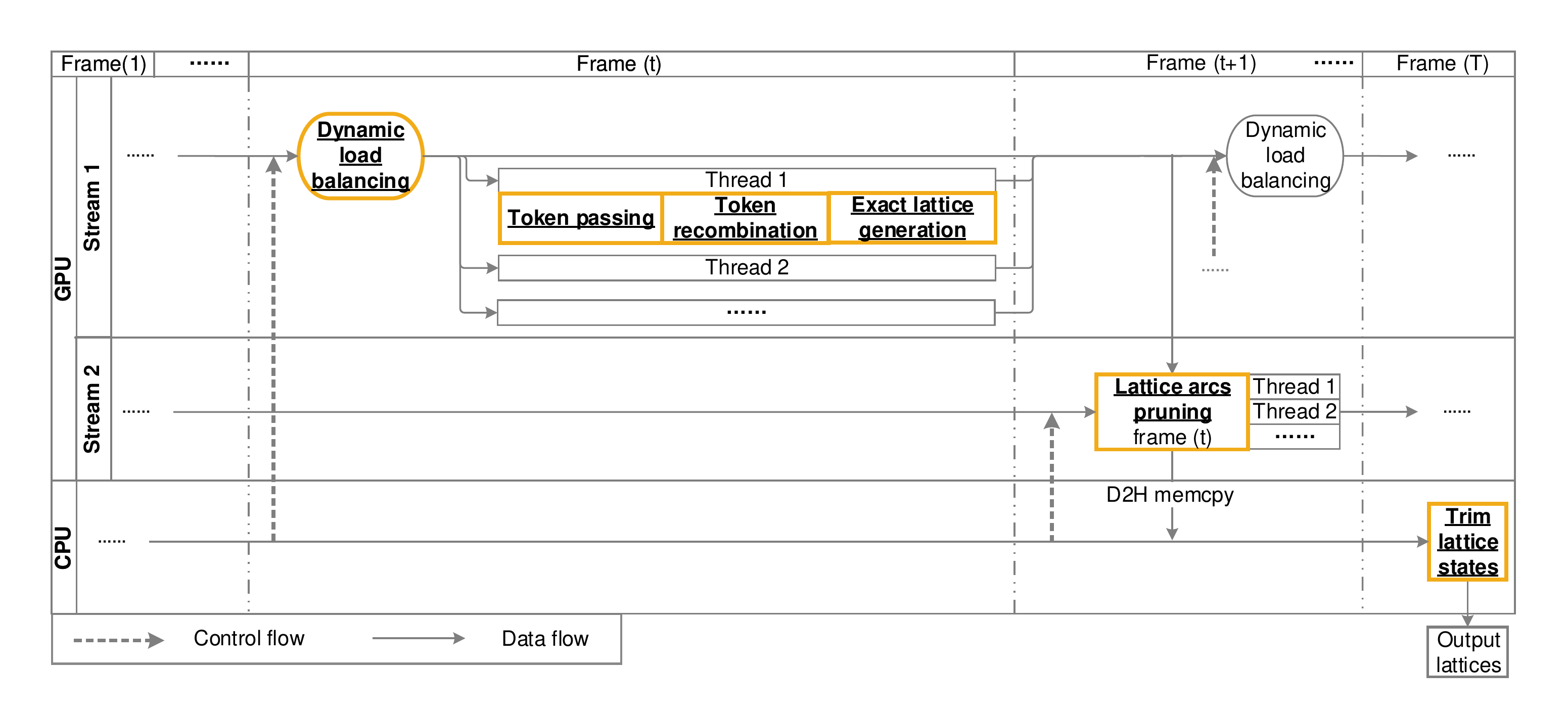}
    \vspace{-3em}
    \caption{\it  Framework of Parallel Viterbi Beam Search with Exact Lattice Processing}
    \vspace{-1em}
    \label{fig:exp-framework}
\end{figure*}
%
Figure~\ref{fig:exp-framework} shows our framework, 
with two GPU concurrent streams performing decoding and lattice-pruning in parallel 
launched by CPU asynchronous calls.
Namely, at frame ($t+1$), stream 2 prunes the lattice generated by stream 1 at frame ($t$).
The procedure of decoding is similar to the CPU version~\cite{povey2011kaldi}
but works in parallel with specific designs discussed in the following sections. 
Load balancing controls the thread parallelism over both WFST states and arcs.
%

\vspace{-0.25em}
\subsection{GPU-based Token Recombination}
\label{sec:atomic}
\begin{figure}[ht]
  \centering
  \vspace{-1.7em}
    \includegraphics[width=1.1\linewidth]{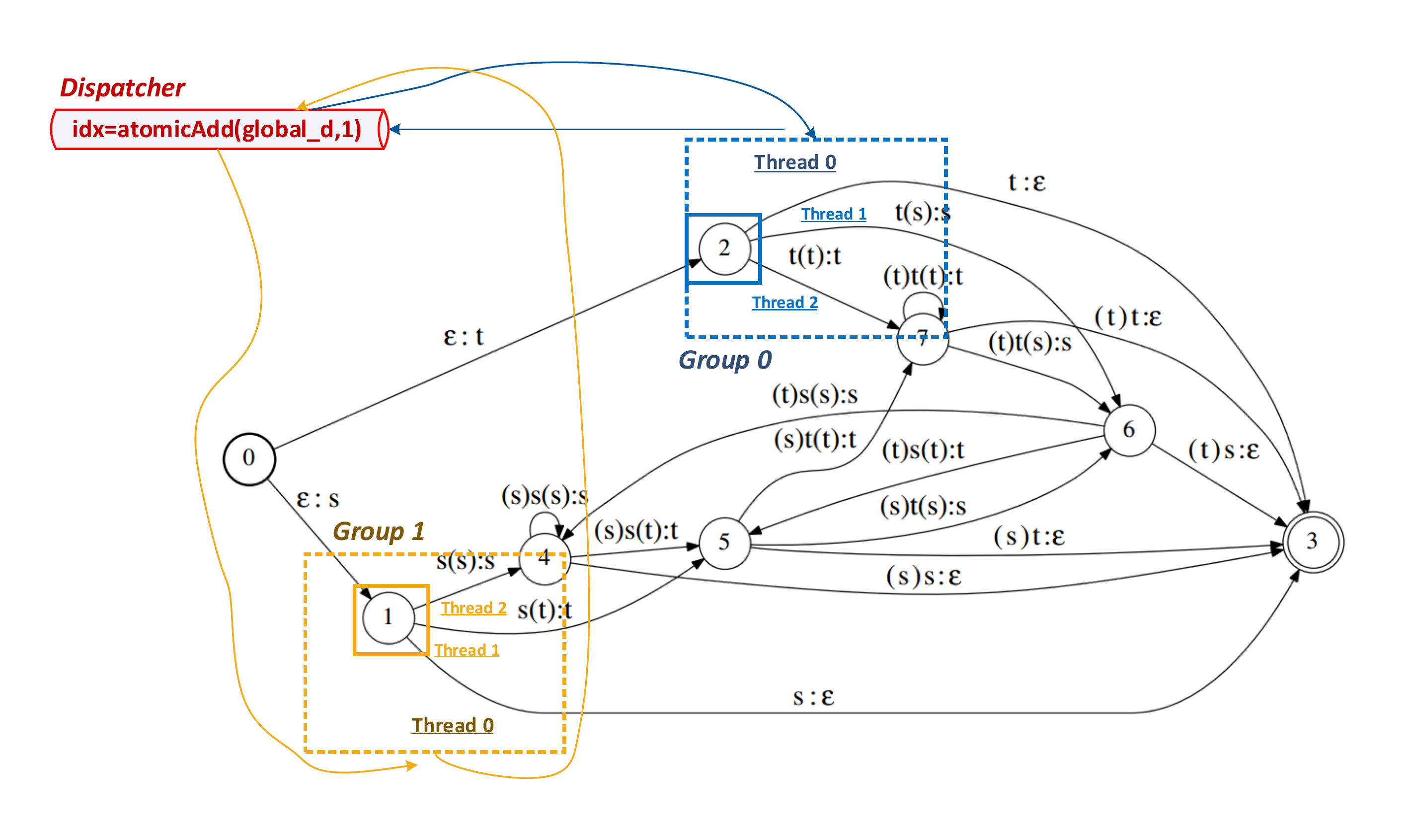}
    \vspace{-2em}
    \caption{\it Example of Dynamic Load Balancing. The dashed box denotes a
      CUDA cooperative group and different groups are with different colors.
        Each group is controlled by Thread 0 of it. After processing all the
        forward links of a state, Thread 0 accesses the dispatcher and the next token  is dynamically decided by atomic operation. Group 0 and 1 work in parallel. }
    \vspace{-1em}
    \label{fig:load-balance}
\end{figure}

\vspace{-0.25em}
The synchronization overhead in  Viterbi search comes from the 
token recombination. 
%
%
In ASR decoding, Viterbi search is reformulated to \textit{token passing algorithm}~\cite{woodland19951994}, where a token represents a partial match up to frame $t$ and each WFST state at frame $t$ holds a single moveable token. 
At each frame, the Viterbi path is obtained  by 
a \textit{token recombination} procedure, where
a \textit{min} operation
is performed on each state over all of its incoming arcs (e.g. state 7 in Figure~\ref{fig:load-balance} and the incoming arcs from state 2, 5 and 7), to compute the best cost and the corresponding predecessor of that state. 

%

\vspace{-0.5em}
\begin{algorithm}[ht]
\caption{Thread-level Token Recombination \textcolor[rgb]{0,0.5,0}{(Inputs: accumulated cost, an out-going WFST arc and a current token)}}
\label{code:atomic}
\begin{algorithmic}[1]
\Procedure{Recombine} {cost, arc, curTok}
\makesavenoteenv{table}
\State oldTokPack = state2tokPack[arc.next\_state]
\State curTokPack = \textit{packFunc}(cost,arc.id) \Comment \textcolor[rgb]{0,0.5,0}{pack into uint64}
\State ret = \textit{atomicMin}~\footnotemark(oldTokPack,curTokPack)
\If { ret $>$ curTokPack }         \Comment \textcolor[rgb]{0,0.5,0}{recombine}
\State  perArcTokBuf[arc.id] = *curTok \Comment \textcolor[rgb]{0,0.5,0}{store token}
\EndIf
\EndProcedure
\end{algorithmic}
\end{algorithm}
\footnotetext{\textit{atomicMin}({\em{*address}}, {\em{val}})~\cite{cuda9}. Computes \textit{min}({\em{*address}}, {\em{val}}), writes the result to {\em{address}}, and returns the original {\em{*address}} .}
\vspace{-0.5em}


The original CPU algorithm conducts recombination in serial.
We discuss how to implement GPU-based token recombination in this section and leave how to parallelize it to the next section.
A naive implementation is by adding critical sections~\cite{lamport1979make}
 to recombine tokens in serial. Such design is inefficient and causes
unexpected deadlocks on pre-Volta GPUs. 
\cite{you2009parallel} performs a per-state reduction on the  token passing results, which requires extra burden of bookkeeping memory to eliminate write conflicts.
\cite{kim2011h} encodes all token data in 32 bits  and does the
recombination by atomic operations. This method loses precision
and makes the decoder dependent on models.

We propose Algorithm~\ref{code:atomic}, which is a general method for GPU token recombination with no precision loss.
The algorithm is performed per GPU thread, and runs in parallel for every arc, e.g. the arc from state 2 to state 7 in Figure~\ref{fig:load-balance} is processed by a thread following this algorithm (details on parallelism are in the following section).
We pack the cost and the arc index into an uint64 to represent
the token before recombination, with the former one in the higher
bits for comparison purpose.
In each frame, we save the token information in an array whose size is the number of arcs.
This ensures there are no write conflicts between threads since each arc can be accessed at most once in each frame.
After passing all tokens, we aggregate survived 
packed tokens, unpack them to get arc indexes, and store token information
from the former array to token data structures  exploiting thread parallelism. 

\vspace{-0.25em}
\subsection{Parallel Token Passing}
\label{sec:para-viterbi}
\vspace{-0.25em}

Another issue in parallel algorithm is load balancing.
%
For each state, we traverse their out-going arcs in parallel, till we reach a final state.
Because WFST states might have different numbers of out-going arcs, the allocation
of states and arcs to threads can result in load imbalance.
Unlike in \cite{you2009parallel,mendis2016parallelizing} where the authors redesigned
the WFST structure, this work does not require any structural change on AM, lexicon or LM.
We propose \textit{static load balancing}, which starts off by assigning roughly
equal number of arcs to each thread, before processing the corresponding arcs.
This design requires the accumulated sum of the arc numbers of
 active tokens~\footnote{A GPU DeviceScan operation over around 10K integers.},
which requires extra computation and kernel launch time.

Motivated by~\cite{alakeel2010}, we also  introduce 
\textit{dynamic load balancing} in Algorithm~\ref{code:load-balance}. 
We use a dispatcher in charge of global scheduling, and make $N$
threads as a group ($N = 32$) to process all arcs from a single
token. When the token is processed, the group requests from the dispatcher a new token.
We implement task dispatching as an atomic operation. Figure~\ref{fig:load-balance}
shows an example.
We compare static and dynamic load balancing in Section~\ref{sec:speedup}.

\vspace{-0.5em}
\begin{algorithm}[ht]
\caption{Grid-level Token Passing \textcolor[rgb]{0,0.5,0}{(N=32; Inputs: the current active token vector)}}
\label{code:load-balance}
\begin{algorithmic}[1]
\Procedure{ Dynamic Load Balancing } {toks}
\State group = cooperative\_groups::tiled\_partition$\left<32\right>$
\If{group.thread\_rank()==0}\Comment\textcolor[rgb]{0,0.5,0}{rank 0 in each group}
\State i = \textit{atomicAdd}(global\_d,1)  \Comment\textcolor[rgb]{0,0.5,0}{allocate new tokens}
\EndIf      
\State i = group.\textit{shfl}(i,0) \Comment \textcolor[rgb]{0,0.5,0}{rank 0 broadcasts i to whole group}
\If {i$>=$sizeof(toks)} return\Comment \textcolor[rgb]{0,0.5,0}{all tokens processed}
\EndIf 
\For {arc in tok2arcs(toks[i])} \Comment \textcolor[rgb]{0,0.5,0}{thread parallelism}
\State {\bf call} \textit{Recombine}(toks[i].cost+arc.cost, arc, toks[i])
\EndFor
\EndProcedure
\end{algorithmic}
\end{algorithm}
\vspace{-1.5em}

\vspace{-0.25em}
\subsection{Exact Lattice Generation}
\label{sec:lat-gen}
An exact lattice~\cite{povey2012generating} is one that stores precise costs
and state level alignments, which is crucial to LM rescoring performance, and enables 
rich post-processing~\footnote{Beside the confidence 
measure~\cite{mangu1999finding} examined in this paper, it also benefits tasks
including minimum Bayes-risk decoding~\cite{goel2000minimum},
system combination~\cite{fiscus1997post}, discriminative training~\cite{povey2005discriminative}, etc. .}. 
Implementing the lattice processing in GPU is non-trivial. 
\cite{kim2014accelerating} simply decodes in GPU and generates lattice in CPU, which significantly slows down the decoder and results in overheads of device-to-host (D2H) memory copy.
We thoroughly solve this problem by redesigning parallel versions of lattice processing algorithms~\cite{povey2012generating} in the following.


In lattice generation, for each token passing operation, a 
lattice arc needs to be stored in GPU memory. We design a GPU vector to 
store arcs using  {\em{push\_back}} function implemented by an atomic 
operation, where the memory is pre-allocated.

\begin{lstlisting}[language=c++,frame=single,morekeywords={int,atomicAdd}]  % Start your code-block

  // implementation of v.push_back(val)
  int idx = atomicAdd(cnt_d, 1); // idx=cnt_d++
  mem_d[idx] = *val;      // store data
\end{lstlisting}

To reduce the overhead of atomic operations, we pre-allocate $K$
vectors and randomly select one to {\em{push\_back}} the lattice
arc into. After token passing, arcs from $K$ vectors are aggregated. 
We process lattice nodes in a similar manner~\footnote{In our 
preliminary experiments, this optimization speeds up lattice 
arcs by $10$ times with $K=32$. The speedup of lattice nodes is small.}.


\vspace{-0.25em}
\subsection{Lattice Pruning}
\vspace{-0.25em}


The parallel lattice pruning is based on the algorithm described in~\cite{ljolje1999efficient,povey2012generating} with necessary modifications for GPU parallelization. 
The original CPU version of lattice pruning iteratively updates extra costs of nodes and arcs until they stop changing, where extra cost is defined as the difference between the best cost including the current arc and the best overall path. 
Arcs with high extra costs are then removed, along with nodes that are not associated with any remaining arcs.
In  GPUs, we i) parallelize the iterative updating of nodes and arcs over GPU threads; 
ii) use a global arc vector to replace the linked lists in the old implementation, for its lack of random access features to enable parallel access; 
iii) implement the extra cost updating as an atomic operation to eliminate write conflicts among threads.

When a lattice arc is pruned, we do not physically remove the arc, as memory allocation is expensive.  Instead, we do a final merging step to aggregate all remaining arcs  using  thread parallelism and the GPU vector proposed in Section~\ref{sec:lat-gen}. Memory consumption of this implementation is discussed in Section~\ref{sec:exp-ana}.
Meanwhile, we do not prune  lattice nodes because: i) we need a static mapping for each arc to trace the previous and the next nodes before and after D2H memory copy. We use frame index $t$ and vector index $i$ to trace a node, thus  node positions in the vector  cannot be changed. ii) the lattice is constructed in CPU by iterating remaining arcs, thus nodes are implicitly pruned. iii)  node D2H copy is done in each frame asynchronously, which does not introduce overheads.


\vspace{-0.50em}
\section{Experiments}
\label{sec:exp}
\vspace{-0.7em}
\subsection{Setup}
\vspace{-0.25em}

In Switchboard 300 hours corpus,
we have two acoustic model baseline setups: one is a ``TDNN-LSTM-C'' model described in ~\cite{peddinti2018low} with lattice-free MMI (LF-MMI) objective~\cite{povey2016purely} with sub-sampled frame rate.
To eliminate the effect of sub-sampling, the other is a stacked bidirectional LSTM network~\cite{sak2014long} with cross entropy  (CE) objective with the original frame rate.

Evaluation is carried out on the NIST 2000 CTS test set. 
A 30k-vocabulary tri-gram LM trained from the transcription of  Switchboard corpus and a Fisher interpolated 4-gram LM are used for  decoding and lattice rescoring respectively. Kaldi 1-best decoder and lattice decoder are taken as CPU baselines~\footnote{{\em{decode-faster-mapped}} and {\em{latgen-faster-mapped}} in Kaldi toolkit.}.

1-best results and lattice quality are both examined. For fair comparisons, lattice densities ({\em{lat.den.}}, measured by arcs/frame)~\cite{woodland19951994} are kept identical for CPU and GPU decoding results.
We report word error rate (WER), lattice rescored WER ({\em{+rescored}}), lattice oracle WER (OWER)~\cite{hoffmeister2006frame} in recognition tasks. Normalized cross entropy (NCE)~\cite{siu1999evaluation} is reported to show the quality of word level confidence measure after lattice rescoring. A decision tree is trained from the posterior histogram  to map the posterior probabilities to confidence scores~\cite{chen2017confidence,chen2017unified}.
Real-time factor (RTF) is taken to evaluate the decoding speed and the speedup factor ($\Delta$) versus corresponding CPU baselines.   
Because of the focus of this work, we report RTFs excluding AM inference in the tables.
Overall RTFs are given when analyzing the tables. As the current goal is to build a fast speech transcription system, latencies of online streaming applications have not been taken into account.
All experiments are conducted on  {\em{E5-2686 v4 @ 2.30GHz}} with 1 socket (8 threads). 
A Tesla V100 is used in default and different architectures from Kepler to Volta are examined in Section~\ref{sec:exp-ana}.

\vspace{-0.25em}
\subsection{Performance and Speedup}
\vspace{-0.25em}
\label{sec:speedup}

Table~\ref{tab:perf-compare} compares the proposed
GPU lattice decoder with the CPU baseline under the same beam.
All indicators are very close if not identical.
The slight differences might be caused by different orders in which states are
visited during decoding.

\begin{table}[thbp!]
\vspace{-0.25em}  
  \caption{\label{tab:perf-compare} {\it  1-best and Lattice Performance (beam=14).  } }
  \vspace{-1em}
  \centerline{
    \begin{tabular}{c c || c c c| c}
      \toprule
      system &  {\em{lat. den.}} &WER & {\em{+rescored}} & OWER & NCE \\
      \midrule
      CPU  & 30.3  & 15.5 & 14.3 & 11.2& 0.322\\
       GPU  & 30.2  & 15.5 & 14.3 & 11.2~\footnote{OWER in this dataset is imprecise as we do not normalize the text in its scoring using {\em{glm}}. In LibriSpeech~\cite{povey2016purely}, we consistently observe that OWER is 23\% of WER (3.4 v.s. 14.7) for both CPU and GPU.  }  & 0.328\\
\bottomrule
    \end{tabular}
  }
  \vspace{-0.25em}  
\end{table}

Decoding RTF speedups of both 1-best and lattice decoders are shown in Table~\ref{tab:speedup}. 
We obtain a 15-fold and 9.7-fold single-sequence speedups for 1-best and lattice decoders respectively.
Most of the  speedups come from parallel traversals of WFST   states and arcs.
The second most significant improvement is from lattice pruning. Beside 
the parallelism benefit similar to that of token passing, the GPU lattice 
pruning reduces the data amount of  D2H memory copy and separate kernel launch costs.
We obtain a 46-fold speedup if sequence parallelism is utilized  with GPU MPS, which reduces context switching of multi-process~\cite{cuda9}.
For comparison, similar multi-process parallelism is applied in CPU but only results in a 1.8-fold 
speedup.

Although the speedup of atomic operation is not significant, it removes the critical section discussed in Section~\ref{sec:atomic}, which enables the use of older GPU architectures, shown in the next section. Meanwhile, dynamic load balancing is a comparable and general solution versus both static load balancing, shown in the 4th row,  and the graph partition in~\cite{mendis2016parallelizing}.

To make a fair comparison of the overall RTFs, we take a separate GPU to do single-sequence inference for both baseline and the GPU decoder. The single-sequence overall RTFs of  baseline and the proposed 1-best decoder are 0.18 and 0.03~\footnote{The AM inference can be further improved by sequence batching~\cite{dixon2009harnessing}, which speeds up 10 times in our preliminary trials.}. Deep integration of inference and decoding can be future topics.

\begin{table}[thbp!]
\vspace{-0.25em}  
  \caption{\label{tab:speedup} {\it  Speedup of the Proposed Method (beam=14). }}
  \vspace{-1em}
  \centerline{
    \begin{tabular}{ m{11em} || c c| c c}
      \toprule
      system  & \multicolumn{2}{c|}{search} & \multicolumn{2}{c}{ + lattice}  \\
       & RTF& $\Delta$&RTF & $\Delta$  \\
      \midrule
      CPU    &  {\bf{0.16}}& {\bf{1.0X}} &{\bf{0.27}} & {\bf{1.0X}} \\
      \ \ + 8-sequence (1 socket)~\footnote{The result is obtained from {\em{latgen-faster-mapped-parallel}} in Kaldi. 1-best decoder do not have such implementation. }  & - & - & 0.15 & 1.8X\\
      \midrule
       GPU &  0.016 & 10X &0.080 & 3.3X \\
       \ \ + atomic operation & 0.015  & 11X &0.077 & 3.5X\\
       \ \ \ \ + dyn. load balancing & {\bf{0.011}}  & {\bf{15}}X & 0.075  & 3.6X\\
       \ \ \ \ \ \ + lattice prune &  - & - & {\bf{0.028}}& {\bf{9.7}}X\\
       \ \ \ \ \ \ \ \ + 8-sequence (MPS) & 0.0035   & 46X &  0.0080 & 34X\\
      \bottomrule
    \end{tabular}
  }
  \vspace{-1em}  
\end{table}

\vspace{-0.25em}
\subsection{Analysis}
\begin{figure}
  \centering
  \vspace{-4em}
    \includegraphics[width=\linewidth]{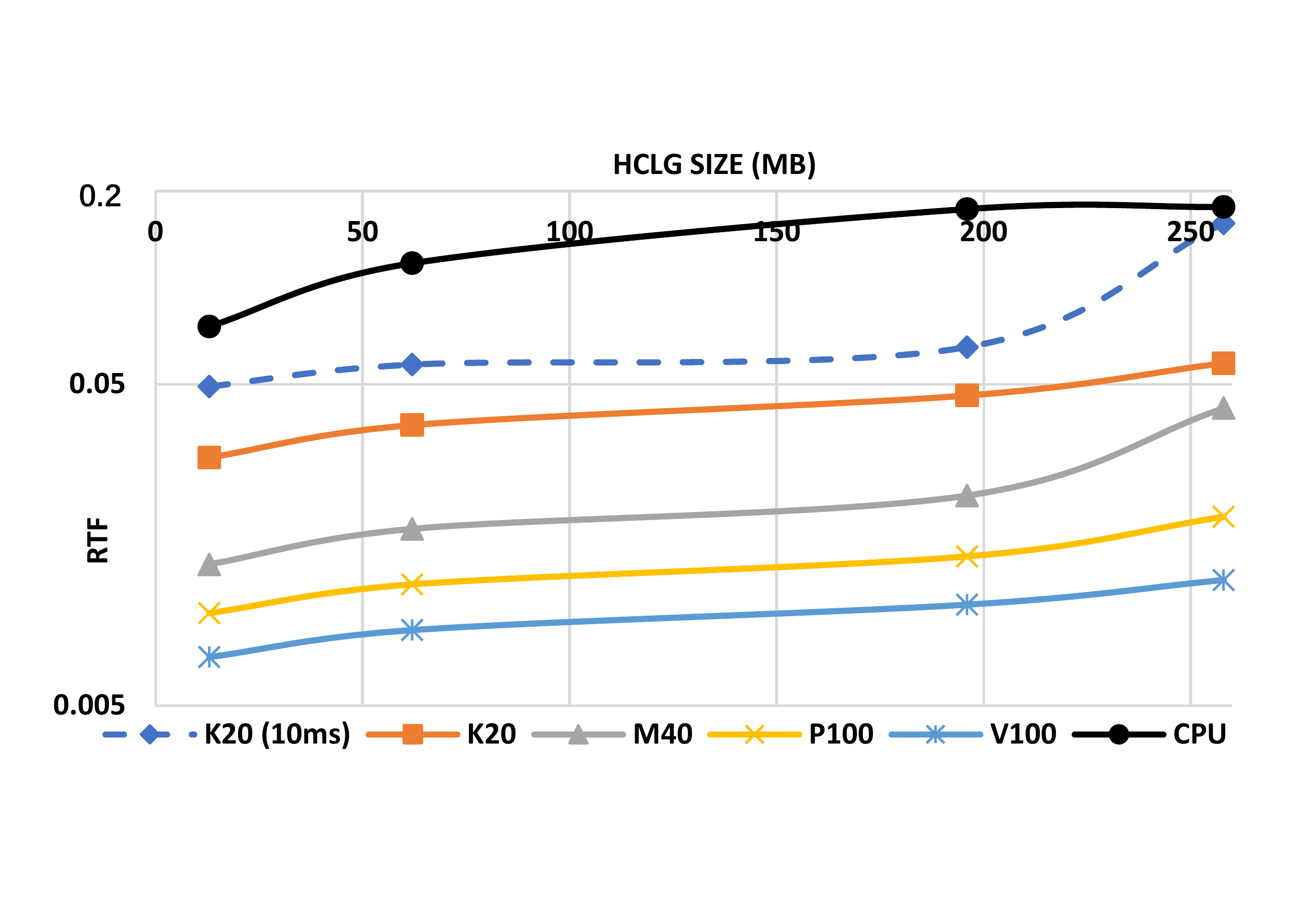}
    \vspace{-4.5em}
    \caption{\it  LM Size, Frame Rates and Architectures Comparison.}
    \vspace{-1.5em}
    \label{fig:exp-ana}
\end{figure}

\vspace{-0.25em}
\label{sec:exp-ana}
Figure~\ref{fig:exp-ana} shows the proposed decoder works well in varieties of ASR systems and consistently improves the speed. 
\begin{itemize} [leftmargin=*]
   \item GPU architectures.
   \vspace{-0.25em}
 \end{itemize} 
 The proposed system can be used in GPU architectures older than Kepler. It achieves a 3-fold speedup in K20 versus CPU. 
 \begin{itemize} [leftmargin=*]
   \item Acoustic model frame rates.
   \vspace{-0.25em}
 \end{itemize} 
 We examine an original frame rate (10 ms) acoustic model, denoted in the dashed line. It works consistently better than the CPU baseline with reduced frame rate. It is worth noting that the 10 ms system is 3 times slower than its reduced frame rate counterpart, as the speech stream is processed in serial. Thus  frame rate reduction~\cite{pundak2016lower,zhc00-chen-tasl2017} is crucial even in GPU decoders.
\begin{itemize} [leftmargin=*]
   \item Language model sizes and memory consumption.
   \vspace{-0.25em}
 \end{itemize} 
The  Fisher interpolated 4-gram LM is pruned to different sizes and compiled to HCLG~\cite{mohri2002weighted} for comparison (13MB, 62MB, 196MB, 258MB). The GPU decoder works consistently better than the CPU baseline. 
Meanwhile, our current implementation decodes at most 11GB WFST in a 12GB TITAN GPU. Details of memory optimization  can be referred to our recent code revision. Although working in a 2-pass framework, the memory consumption can be further optimized in future.
Moreover, CPU speed changes slightly with WFST sizes from 196MB to 258MB. Because of per-transition beam pruning, the number of tokens   is not linear with WFST size. Our GPU decoder always passes tokens through arcs in parallel and does not have the phenomenon. Better scheduling strategies can be future work.



\begin{figure}[ht]
  \centering
  \vspace{-3em}
    \includegraphics[width=1\linewidth]{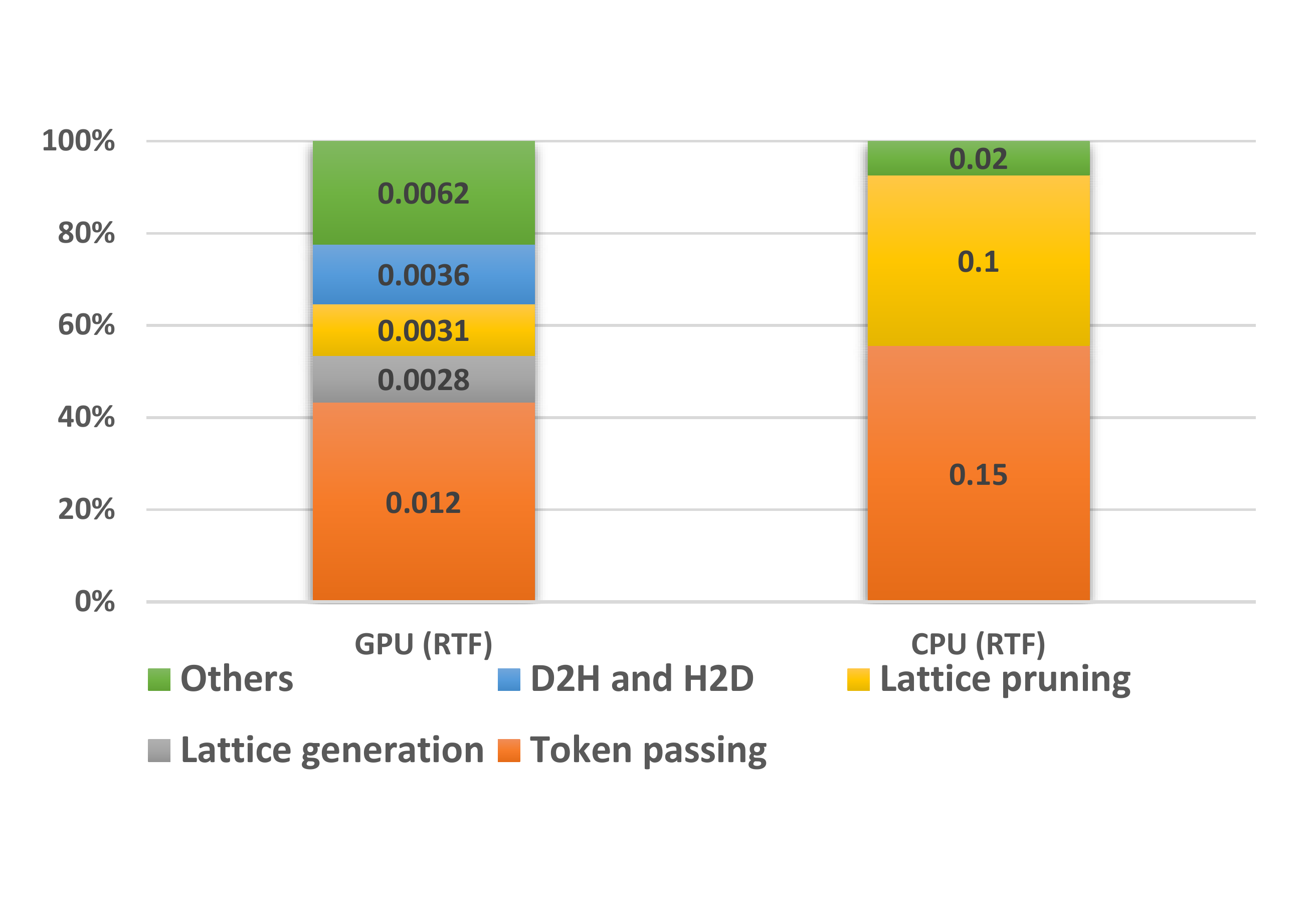}
    \vspace{-4.5em}
    \caption{{\it Profiling of the Decoders (best viewed in color).} }
    \vspace{-1em}
    \label{fig:exp-profile}
\end{figure}

\begin{itemize} [leftmargin=*]
   \item Profiling.
   \vspace{-0.25em}
 \end{itemize} 
  Figure~\ref{fig:exp-profile} 
 compares the profiling result of the baseline CPU decoder~\footnote{Lattice generation statistics in CPU is included in token passing.} and  our final system.
  The GPU decoder significantly speeds up both token passing and lattice processing, which take up most of the decoding time in CPU. Meanwhile, the ``other'' part in GPU includes kernel launch time and some  synchronization pending. Optimizing them can be future topics. 

\vspace{-0.25em}
\section{Conclusions}
\vspace{-0.25em}
\label{sec:conclu}

In this work, we describe initial work on an extension of the Kaldi toolkit that supports  WFST decoding on GPUs. We design parallel versions of decoding and lattice processing algorithms. The proposed system significantly and consistently speeds up  the CPU counterparts in a varieties of architectures. 

The implementation of this work is open-sourced. Future works include the deep integration with sequence parallel acoustic inference in GPUs.

\bibliographystyle{IEEEtran}

\bibliography{mybib}


\end{document}